\documentclass[a4paper]{article}

\usepackage{INTERSPEECH2020}
\usepackage{amsmath}
\usepackage{paralist}
\usepackage{tabularx}
\usepackage{leftidx}

\title{Speech-Image Semantic Alignment Does Not Depend on Any Prior Classification Tasks}
\name{Masood S. Mortazavi$^1$}
\address{
  $^1$Futurewei Technologies, Santa Clara, California, USA}
\email{masood.mortazavi@futurewei.com}

\begin{document}
\maketitle 
\begin{abstract}
Semantically-aligned $(speech, image)$ datasets can be used to explore ``visually-grounded speech''. 
In a majority of existing investigations, features of an image signal are extracted using neural networks ``pre-trained'' on other tasks (e.g., classification on ImageNet). In still others, pre-trained networks are used to extract audio features prior to semantic embedding. 
Without ``transfer learning'' through pre-trained initialization or pre-trained feature extraction, previous results have tended to show low rates of recall in $speech \rightarrow image$ and $image \rightarrow speech$ queries.
%
%

Choosing appropriate neural architectures for encoders in the speech and image branches and using large datasets, one can obtain competitive recall rates without any reliance on any pre-trained initialization or feature extraction: $(speech,image)$ semantic alignment and $speech \rightarrow image$ and $image \rightarrow speech$ retrieval are canonical tasks worthy of independent investigation of their own and allow one to explore other questions---e.g., the size of the audio embedder can be reduced significantly with little loss of recall rates in $speech \rightarrow image$ and $image \rightarrow speech$ queries. 

%

\end{abstract}
\section{Introduction}
The speech-image alignment has long fascinated philosophers and poets \cite{rylemind, poeticimage}. We can imagine (``image'') what we hear. 

The correlations of textual descriptions and images, including image recall and image generation through textual description has been explored in \cite{feifei2015, vinyal2015, attend2015, reed2016, attnGAN}. Yet text is a relatively new invention in the human experiential repertoire. 
In fact, babies first grasp language and then use it in the real multi-modal world involving other senses such as vision and touch\cite{dupoux2018, spelke1990, h2020iclr}. 
Therefore, one expects to find a semantic correlation in multi-modal (e.g., auditory and visual) experience without any ``ground-truth,'' annotation or labels involved.
In \cite{s2014nips, h2015arsu, h2016nips}, ``visually-grounded speech'' began to be investigated and the properties of multi-modal semantic embedding of speech and of images were explored. 
Later research has investigated how near-final layers of networks producing latent semantic embedding (for speech-image semantic alignment) can also be used to extract feature-sets relevant to speech segmentation\cite{h2017acl, c2017acl, h2018lncs, m2019is, h2020iclr}. Other works have continued to explore these and various other aspects of visually-grounded speech \cite{k2017is,h2019ic,r2018ic, k2019au, a2019is}. An excellent, far more complete review of recent work can be found in \cite{h2020iclr}.

This paper returns to the original question of recall in $speech \rightarrow image$ and $image \rightarrow speech$ queries. 
Recall rates are not merely an effectiveness measure for semantically-aligned multi-modal embedding but they are also significant on their own because in real life we recall images through the hearing and the expression of speech. 

In most earlier work, which have reported recall rates on $speech \rightarrow image$ and $image \rightarrow speech$ queries (e.g., those reviewed above), transfer learning based on supervised tasks involving \textit{annotated} data---such as visual classification on ImageNet---has been used.  
Similarly, \cite{m2019is} has noted that while the use of pre-extracted audio features such as Multilingual Bottleneck (MBN, \cite{n2014ICASSP, r2017csl}) may be justifiable in low-resource and small-dataset settings, ``learned audio features face the same issue as word embeddings, [because] humans learn to extract useful features from the audio signal as a result of learning to understand language and not as a separate process.'' In short, MFCC can be computed for any speech signal without needing any other data while MBN features are learned on top of MFCC to recognize \textit{annotated} phoneme states. 

In this paper, the possibility of good recall in the absence of any transfer learning has been explored. The embedders used begin with random initialization and train directly using an image's pixels and an audio clip's MFCC frame sequence as direct input to the trainable neural networks. The results presented will demonstrate that no pre-training (on annotated data) and no pre-trained feature extraction are required in order to arrive at competitive levels of recall in $speech \rightarrow image$ and $image \rightarrow speech$ queries. 

The main contributions of this paper are as follows:
\begin{inparaenum}[(1)]
\item
A demonstration that, with adequate data, neither pre-trained network initialization nor pre-trained feature extraction are necessary when looking to obtain competitive recall rates in $speech \rightarrow image$ and $image \rightarrow speech$ queries.
\item
A novel combination of embedders for semantically aligned $(speech,image)$ pairs which produce state-of-the-art recall rates \textit{in the absence of any pre-training or pre-trained feature extraction}. All versions use raw inputs of pixels and MFCCs as inputs to semantic embedders.  
\item 
A study of how the number of bidirectional recurrent layers in the speech embedder and the size of latent space impact recall rates: One can significantly reduce the number of recurrent layers (the size of the audio embedder with respect to the size of the image embedder) and still retain competitive recall rates in $speech \rightarrow image$ and $image \rightarrow speech$ queries. 

Many questions (e.g., regarding the impact of network size) are impractical to pose precisely when transfer learning has been used based on \textit{a priori} tasks. 
\end{inparaenum}
\section{Speech-Image Co-Embedding}
\begin{figure}[t]
\small
  \centering
  \includegraphics[width=0.75\linewidth]{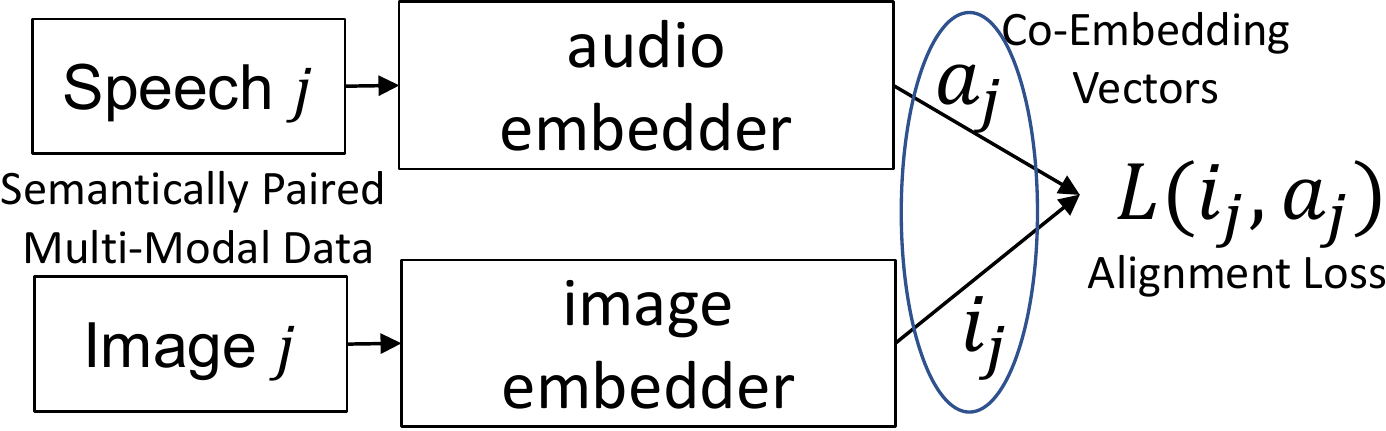}
  \caption{
\footnotesize
Schematic diagram of encoding branches for multi-modal alignment.}
  \label{fig:branches}
\end{figure}

Here, the main objective is to evaluate and demonstrate that the task of $speech \rightarrow image$ and $image \rightarrow speech$ retrieval, by itself, is a fundamental task that can be used to train embedders which capture multi-modal semantic alignment. 
As such, the study and results presented here do not utilize or depend on any label-driven tasks such as training on ImageNet to extract features or to initialize networks. Training of the embedding neural networks starts with pixels and MFCC data for semantically aligned $(speech, image)$ pairs and with randomly initialized networks. 
Figure \ref{fig:branches} shows the dual-branch compute graph which embeds the semantically aligned pair into a common latent space and evaluates the alignment objective function. 
\section{Datasets}
Several semantically aligned $(speech, image)$ datasets exist. This work considers two  popular sets: ``Flicker8k'' (``F8K'') developed in \cite{f8k, h2015arsu}, and ``MIT Places 205'' (``P205'') whose image component is documented in \cite{p205} and whose audio component was developed in \cite{h2016nips, h2017acl, h2018lncs}. 
Of the two aforementioned datasets, F8K is much smaller. It has only 8000 images which are  divided into three sets: training (6000), validation (1000) and test (1000). F8K contains 5 audio clips per image. By contrast, P205 has English audio captions spoken by 2,683 unique speakers for  
402889 of the images in an even larger ``Places 205'' image set\cite{p205}. 
F8K is a very small dataset: The networks reported in table \ref{tab:netsize} (specially the larger ones) overfit its training set as measured by R@10 values of 87\% to 98\%.
As such, F8K is used only for validating the proposed methods, network implementations and training schemes while P205 is used for extensive experiments.  
%
%
\section{Embedders}
This section provides a summary of the architectural designs for the speech and image embedders used here. 
\begin{table}[th]
\addtolength{\tabcolsep}{-3pt}
\footnotesize
  \caption{
\footnotesize
Network sizes, given as the number of trainable float-32 parameters in each of the three embedding network pairs discussed in this paper.}
  \label{tab:netsize}
  \centering
  \begin{tabular}{lrrcr}
    \toprule
    \multicolumn{1}{c}{
    \centering \begin{tabular}{c} Embedder \\ Pair Name\end{tabular} } & 
    \multicolumn{1}{c}{
    \centering \begin{tabular}{c} Audio \\ Embedder\end{tabular} } &
    \multicolumn{1}{c}{
    \centering \begin{tabular}{c} Image \\ Embedder\end{tabular} } &
    \multicolumn{1}{c}{
    \centering \begin{tabular}{c} GRU \\ Layers\end{tabular} }&
    \multicolumn{1}{c}{
    \centering $N$}
    \\
    \midrule
    DG4A2048 & 63,898,816 & 36,155,776 & 4 & 2048\\
    \midrule
    DG4A1024 & 16,228,544 & 33,402,240 & 4 & 1024\\
    \midrule
    DG3A1024 & 11,503,808 & 33,402,240 & 3 & 1024\\
    \midrule
    DG2A2048 & 26,125,504 & 36,155,776 & 2 & 2048\\
    \midrule
    DG2A1024 & 6,779,072  & 33,402,240 & 2 & 1024\\
    \bottomrule
  \end{tabular}  
\end{table}
\subsection{Speech Embedder}
Some previous investigations with the P205 dataset (e.g., \cite{h2020iclr,h2016nips,h2018lncs}) have used Convolutional Neural Networks (CNNs) to encode audio captions in P205, with all speech MFCC frame sequences trimmed at a constant length. 
In \cite{m2019is}, a novel structure for embedding speech was proposed. This latter architecture has been adopted here: The speech embedding network starts with a $1$-dimensional convolution with $64$ kernels of length $6$, computing convolutions along the input (a time-ordered MFCC sequence), followed by one or more bidirectional Gated Recurrent Unit (GRU) layers. These Bi-GRU layers all have the same configuration: hidden size equals $N/2$, where $N$ is the dimension of the latent semantic vector space. The hidden state (in the two directions of the final Bi-GRU layer) are concatenated at each position and sent to an attention layer as described in \cite{m2019is}, where: 
\begin{equation}
{\alpha}_t = softmax(V tanh(W h_t + b_w) + b_v)
\label{eq:coeff}
\end{equation}
\begin{equation}
Att(h_1, ..., h_t) =\sum{ {\alpha}_t \circ h_t }
\label{eq:attn}
\end{equation}
Here, the $h_t$ are concatenated internal states of the last forward and backward components of the Bi-GRU at position $t$. They have $N$ elements. $W$ has an internal size of 128 and $V$ has an internal size of $N$ as in \cite{m2019is} and $b_w$ and $b_v$ are bias factors.
The output of this single-head attention layer---i.e., $Att(h_1, ..., h_t$) which is of size $N$---is normalized to provide the $N$-dimensional unit embedding of the audio caption.   
\subsection{Image Embedder}
\label{sec:image_embedder}
Image embedders used in previous works have been of two variety when it comes to pre-training by image classification: Either features in final layers of pre-trained networks (e.g., VGG-16 on ImageNet) have been extracted from images and used in some trainable  layers to embed the image in a common semantic latent space (e.g., as in \cite{m2019is,h2016nips}) or such pre-training has been used as an initialization scheme (e.g., as in \cite{h2020iclr,h2018lncs}). 

The present study uses DenseNet \cite{densenet} to embed images in the common semantic latent space with neither pre-trained initialization nor feature extraction involved. 
DenseNet is equipped with dense blocks that have proven to be compact feature extractors alleviating the ``vanishing gradient'' problem using dense skip connections that allow feature reuse and induce back-propagation short-cuts. The reader is referred to table 2 in \cite{densenet} for comparisons of DensNet with other common network architectures used in computer vision. 

The DenseNet image embedder used here is almost identical to DenseNet-264 in \cite{densenet}: It has a \textit{growth rate} of 32, starts with a Conv(7×7) and a \textit{block configuration} of $(6, 12, 64, 48)$. The growth rate represents the number of feature maps produced by each of the convolutional layers within a dense block. The configuration gives the number of convolutional layers within each block. Thus, DenseNet-264 has 6, 12, 64, and 48 convolutional layers in each of its four block. These layers are each composed of a BN-ReLU-Conv(1×1)-BN-ReLU-Conv(3×3) combination.  
The final tiers of the DenseNet embedder used here consists of an average pooling operation over the whole plane of the feature maps and then a linear layer that produces a vector of size $N$. 
In this study, two values were adopted for the latent space dimension $N$: 2048 and 1024. 
\subsection{Objective Function}
\label{subsec:objective}
``Hinge loss'' (``triple loss'' or ``max margin objective'' \cite{feifei2015}) has been used to compute the misalignment of a semantic embedding pair $(a,i)$ for a semantically-aligned data pair $(speech, image)$ relative to semantic embeddings for other data items in a mini-batch. This loss function can be computed as follows: 
\begin{multline}
L = \frac{1}{M} \sum_{(a,i),(a^\prime,i^\prime) \in B} 
max(0, S(a, i^\prime) - S(a, i) + \beta) \\
+ max(0, S(i, a^\prime) - S(i, a) + \beta) \\
\label{eq:hinge}
\end{multline}
Here, $B$ represent the mini-batch set, $M$ represents 
$\vert B \vert \times \vert B-1 \vert$, 
and similarity function $S$ is computed as $S(a,b)={a}\cdot{b}$, where vectors are normalized. Hinge threshold $\beta$ is a small value, and for all the results here, it has been set to $0.2$. 
While some have used importance sampling to sub-sample $(a^\prime,i^\prime) \neq (a,i)$ , and others have used the maximum non-aligned similarities, the results shown in this paper use all non-aligned pairs $(a, i^\prime)$ and $(a^\prime, i)$ in the batch, derived from all  $(a^\prime,i^\prime) \neq (a,i)$ in the batch.
%
\section{Training and Testing of Embedder Networks}
Each pair of embedder networks, one for speech and one for images, were trained together using the same P205 set of 400K semantically aligned $(speech,image)$ data items.
Audio sample rate used was 16,000.
\subsection{Preparation of MFCC files}
To accelerate training, the MFCC tensors were pre-computed for all the audio files and placed in corresponding files in a parallel directory structure. MFCC parameters were the default values in TorchAudio's transformer, using 40 coefficients and FFT size of 400, which at the given sample rate produces frame sizes of 25 msec, and frame shifts of 12.5 msec. 
Any pair of $(speech, image)$ data whose corresponding MFCC tensor involved more than 8192 frames was dropped.  This threshold policy excluded less than 32 data items from the combined set of 402889 in P205. (Only 1 data item was excluded from the test set.) This  threshold was adopted in order to manage memory pressure on the GPU devices.    
\subsection{Optimizer and Scheduler}
\label{subsec:optim}
Adam optimizer \cite{k2014adam} was used and experiments were conducted with various kinds of learning rate (LR) schedulers, e.g., cosine-annealing LR scheduler (CALR) and CALR with pre-scheduled warm restarts (CALWR), both proposed in \cite{calr} and as implemented in PyTorch\cite{pytorch}, and cyclic learning rate scheduler (CYLR) proposed in \cite{cylr} and as implemented in \cite{m2019is}. 

Using the F8K dataset to explore the trade-offs between CALR and CYLR schedulers, it was found that CALR produced the tightest function approximation. 
In 200 epochs, subject to other variations, CALR scheduler produced R@10 values ranging from 80 to 98\% on the training set. CYLR, produced R@10 values in the 70\% range. Following this observation, the CALR and CALWR schedulers were adopted for all experiments on the much larger P205 dataset.

\begin{figure}[t]
\small
  \centering
  \caption{
\footnotesize
R@10 convergence in a typical training run. This run is for DG2A2048. Encoding networks were check-pointed at the end of each epoch. Query tests were performed separately after the completion of the full training run which took about 14 days with PyTorch\cite{pytorch} running on two NVIDIA v100 GPUs.}
  \includegraphics[width=0.75\linewidth]{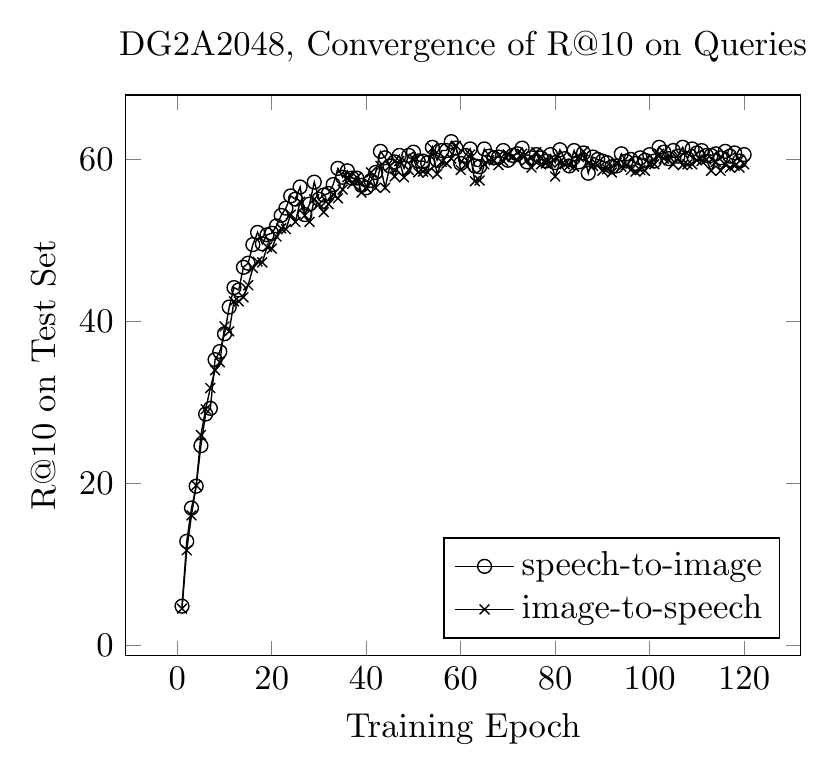}
  \label{fig:convergence}
\end{figure}

As shown in \cite{calr}, CALR scheduler allows for repeated warm restarts. It was observed that using warm restarts either once or twice, over short training runs starting from the best performing checkpoints of an initial run, could improve the performance of R@10 between 1.5\% to 10\%.   
A smaller number of epochs in the initial run of a warm restart is recommended in CALR\cite{calr}. This recommended scheduling procedure had to be modified in some cases due to the practicalities of using a shared set of devices. 
For example, DG2A2048 pair of encoders were first trained up to 120 epoch (see figure \ref{fig:convergence} for details), the 59th epoch's networks were chosen as the best embedders and were then trained for an additional 30 epochs. The initial learning rate on the second run were \textit{reset to half} the initial learning rate of the first run. The second training run produced roughly a gain of 1.5\% in R@10. In the case of DG4A2048, whose audio branch is much larger, a first run of 50 epochs was followed by a second warm run of 10 epochs. This produced a R@10 gain of 10\%. Following the more strict warm restart schedule, as proposed in \cite{calr}, the training runs for networks DG3A1024 and DG4A1024 used an initial epoch length of $T_0=1$ and a restart multiplier of $2$, leading to warm restarts involving 1, 3, 7, 15, 31, etc., epochs.    
Networks were trained using minibatch sizes of 16 to 64, and some for as long as 127 epochs, including warm restarts. The results shown in table \ref{tab:R@10} use a mini-batch size of 64. Larger mini-batches tend to produce better results for the semantic alignment task here. A larger mini-batch can also stand in for a reduced initial learning rates in CALR.\footnote{This learning rate scheduling strategy as given in \cite{batch_size} might be worth-while to try, along with CALR or CALWR\cite{calr} in future studies.} 
\subsection{Size Variations in Embedder Pairs}
Experiments were conducted with five speech-image encoder pairs whose sizes (as number of trainable parameters) are given in table \ref{tab:netsize}.
The largest networks are in the DG4A2048 encoder pair. Its audio branch has 4 Bi-GRU layers (``G4'') followed by a single-head self-attention layer (``A''). Its audio and image branches each produce a vector encoding of length ``2048''. DG4A2048's speech encoder has 64 million parameters.
Next, DG2A2048's audio encoder has 2 Bi-GRU layers and a total of 26 million parameters. 
Its image encoder is the same as DG4A2048. 
The other three pairs, DG2A1024, DG3A1024 and DG41024, produce embedding vectors of length 1024, and their audio branches include 2, 3 and 4 Bi-GRU layers and a total of 6.8, 11.5 and 16.2 million parameters.  
In all five cases, the image-embedding DenseNet (``D'') has a block configuration as described in section \ref{sec:image_embedder}. 
In training, it consumes a random $224 \times 224$ crop of the P205 ($256 \times 256$) images randomly flipped horizontally with a probability of $0.5$.
\subsection{Testing}
The test set consists of 1000 $(speech, image)$ pairs, as described in \cite{h2020iclr, h2018lncs} and indicated in P205 meta-data. Only one pair, with a very long MFCC sequence, was excluded from the original test set. Its exclusion relieves GPU memory pressure that would occur on mini-batches containing it. 
%
To evaluate, all the speech encodings and all the image encodings and then the top-10 recall rates (R@10) in the $speech \rightarrow image$ and $image \rightarrow speech$ queries are computed on the test set.
For example, figure \ref{fig:convergence} provides a plot of R@10 evaluations on the test set for the network pairs check-pointed at the end of each epoch in a typical training run. 
\section{Results}
Table \ref{tab:R@10} lists earlier results and those obtained using the five embedder pairs in this study. 
All results on P205 dataset (shown in table \ref{tab:R@10}) have used MFCC as direct input to the audio embedder.

Only the results in bold use P205(400K) dataset \textit{but} \textit{neither} any pre-trained initialization \textit{nor} any features extracted using networks trained on any other \textit{a priori} tasks such as image classification on ImageNet. 
In other words, they involve \textit{no transfer learning} and are directly comparable.  

\begin{table}[th]
\addtolength{\tabcolsep}{-3pt}
\footnotesize
\caption{
\footnotesize
Recall in $speech \rightarrow image$ and $image \rightarrow speech$ queries.
Column ``TL'' characterizes any transfer learning used to obtain a given result.  
With $x$ standing for any non-repeating combination of 
of $i$ (image) or $a$ (audio), $\protect\leftidx{_x}{P}{}$ means pre-trained initialization of  branch or branches $x$ (or all but a few layers of $x$) on some prior (mostly, classification) task, and $\protect\leftidx{}{P}{_x}$ means the dependence of branch or branches $x$ on features extracted by a network trained on some prior (mostly, classification) task.
Note that $\protect\leftidx{}{P}{_x}$ involves a stronger version of transfer learning than $\protect\leftidx{_x}{P}{}$.
The R@10 values are written as pairs with the first value corresponding to ``$speech \rightarrow image$'' and the second ``$image \rightarrow speech$'' queries.  
The table also contains the list of networks forming the core of the image encoder ($I_e$). There is a much larger variation in the speech/audio encoder, e.g., CNNs, GRUs and RHUs, and these have not been listed, here. 
}
  \label{tab:R@10}
  \centering
  \begin{tabular}{llcrr}
    \toprule
    \multicolumn{1}{c}{
    \textbf{Source}} & 
    \multicolumn{1}{c}{
    \textbf{Dataset}} &
    \multicolumn{1}{c}{
    \textbf{TL}} & 
    \multicolumn{1}{c}{
    \textbf{$I_e$}} & 
    \multicolumn{1}{c}{
    \textbf{R@10}}
    \\
    \midrule
    \cite{h2020iclr} 
    & $P205$(400k)  & \textit{$\leftidx{_i}{P}{}$} & Resnet50 & $0.735$~~~\\
    \cite{h2018lncs}MISA 
    & $P205$(400k) & \textit{$\leftidx{_i}{P}{}$} & VGG16  & $0.604$ / $0.528$~~~\\
    \cite{h2016nips,h2018lncs}      
    & $P205$(120k) & \textit{$\leftidx{}{P}{_i}$} & VGG16 & $0.548$ / $0.463$~~~\\
    \cite{h2017acl,h2018lncs}
    & $P205$(200k)& \textit{$\leftidx{}{P}{_i}$} & VGG16& $0.564$ / $0.542$~~~\\
    \cite{h2017acl}
    & $P205$(200k)& \textit{$\leftidx{}{P}{_i}$} & VGG16& $0.431$ / $0.438$~~~\\
    \cite{h2015arsu} & $F8K$(40k/8k)  & \textit{$\leftidx{}{P}{_{ia}}$} & RCNN & $0.179$ / $0.243$~~~\\
    \cite{m2019is} & $F8K$(40k/8k)& \textit{$\leftidx{}{P}{_{ia}}$} & Resnet152 & $0.376$ / $0.452$~~~\\
    \cite{m2019is} & $F8K$(40k/8k)& \textit{$\leftidx{}{P}{_{ia}}$} & Resnet152 & $0.485$ / $0.561$~~~\\
    \cite{c2017acl} & $F8K$(40k/8k)& \textit{$\leftidx{}{P}{_{i}}$} & VGG16 & $0.253$ /~~~\\
    \cite{g2019large} & $F8K$(40k/8k)  & \textit{$\leftidx{_{ia}}{P}{}$} & IncResNetV2 & $0.495$ / $0.558$~~~\\
    \cite{g2019large} & $F8K$(40k/8k)  & \textit{$\leftidx{_{a}}{P}{}$} & IncResNetV2 & $0.211$ / $0.241$~~~\\
    \cite{g2019large} & $F8K$(40k/8k)  & \textit{$\leftidx{_{i}}{P}{}$} & IncResNetV2 & $0.279$ / $0.352$~~~\\
    \midrule
    \midrule
    \cite{g2019large} & $F8K$(40k/8k)  & No & IncResNetV2 & $0.101$ / $0.124$~~~\\
    \textbf{\cite{h2018lncs}MISA} & $P205$(400k)  & No & VGG16 & $0.314$ / $0.291$~~~\\
    \cite{h2016nips}Mean & $P205$(120k)& No & VGG16 & $0.299$ / $0.295$~~~\\
    \cite{h2016nips}Max  & $P205$(120k)& No & VGG16 & $0.309$ / $0.291$~~~\\
    \midrule
    \textbf{DG4A2048} & $P205$(400k)  & No & DenseNet & $0.620$ / $0.622$~~~\\
    \textbf{DG4A1024} & $P205$(400k)  & No & DenseNet & $0.652$ / $0.594$~~~\\
    \textbf{DG3A1024} & $P205$(400k)  & No & DenseNet & $0.677$ / $0.671$~~~\\
    \textbf{DG2A2048} & $P205$(400k)  & No & DenseNet & $0.628$ / $0.626$~~~\\
    \textbf{DG2A1024} & $P205$(400k)  & No & DenseNet & $0.590$ / $0.581$~~~\\
    \bottomrule
  \end{tabular}
\end{table}
%
%
%
\section{Discussion and Conclusion}
The results obtained in the current study make a drastic improvement over earlier results on P205(400K) \textit{with no transfer learning involved}, i.e. R@10 values of $0.677$ / $0.671$, for DG3A1024, compared to $0.314$ / $0.291$, for \cite{h2018lncs}MISA. 
These results can probably be further improved with hyper-parameter and architectural adjustments. 
%

Note that the best, alternative semantic alignment result on this dataset (i.e., \cite{h2020iclr}) uses a \textit{pre-trained} Resnet-50, from \cite{resnet}, on the image branch with $N=1024$. It is not directly comparable in the none-pre-trained setting here.  

In the non-pre-trained setting, one can also pose questions regarding the relative size of embedding networks. 
The last five rows of results in table \ref{tab:R@10} show that within the Bi-GRU architectural family, a range of audio-embedder network sizes (see table \ref{tab:netsize}) may still produce competitive recall rates in the absence of any form of pre-training on any \textit{a priori} tasks. 
Mid-sized DG3A1024 produces better test evaluation results than network pairs that are much larger or much smaller. Larger networks overfit the training set and the smaller ones do not have adequate capacity. 
The network pair with the smallest audio embedder, DG2A1024, still does comparatively well, with R@10 values of $0.590$ / $0.581$.  
Table \ref{tab:R@10} also shows that the proper choice of the image embedder, $I_e$, can  make a significant difference to $(speech, image)$ semantic alignment.

%
%
In summary, one can conclude that given 
\begin{inparaenum}[(a)] 
\item no pre-trained initialization or feature extraction, 
\item use of a Bi-GRU architecture on the audio branch, and 
\item use of DenseNet architecture on the image branch, one can produce well-aligned semantic embedding of images and their spoken description. Some of the strength of semantic alignment shown in the results here may also be attributable to 
\item the CALR/CALWR optimization scheduler.
\end{inparaenum}
These combined choices, have produced competitive results that require neither network initialization nor feature extraction based on other tasks such as image classification or phoneme recognition. 
The only objective on which any of these networks have trained is the semantic alignment of images and speech (audio captions). 

All this indicates that the $(speech, image)$ semantic alignment can be viewed as a canonical, independent task of its own. 
%

%
The results reported here show that using a large dataset of semantically aligned $(speech, image)$ pairs, competitive rates of recall, R@10, can be obtained in the absence of any transfer learning.  
They also show that we can reduce the size (i.e., the number of parameters) of the speech embedder  significantly in comparison to the image embedder with very little reduction in recall rates in $speech \rightarrow image$ and $image \rightarrow speech$ queries. 

In future work, it would be interesting 
\begin{inparaenum}[(a)] 
\item
to investigate whether comparable reductions can occur on the image branch, 
\item
to explore the relation between the two branches; e.g., the relative network capacity in each branch, 
\item
to study the temporal and spatial alignment of extracted audio and image features, and
\item 
to see whether down-stream tasks can use features extracted by embedders trained on $(speech, image)$ semantic alignment task alone.  
\end{inparaenum} 
 
%

\bibliographystyle{IEEEtran}

\bibliography{s2ibib}

\end{document}